%% file: main.tex
\documentclass[10pt,twocolumn,letterpaper]{article}

\usepackage{wacv}
\usepackage{times}
\usepackage{epsfig}
\usepackage{graphicx}
\usepackage{amsmath}
\usepackage{amssymb}
\usepackage{customization}

\makeatletter
\@namedef{ver@everyshi.sty}{}
\makeatother
\usepackage{tikz,pgfplots}
\pgfplotsset{compat=1.16}

\usepackage[accsupp]{axessibility}  

\def\wacvPaperID{5} 

\wacvfinalcopy 

\ifwacvfinal
\def\assignedStartPage{1} 
\fi

\ifwacvfinal
\usepackage[breaklinks=true,colorlinks,bookmarks=false]{hyperref}
\else
\usepackage[pagebackref=true,breaklinks=true,colorlinks,bookmarks=false]{hyperref}
\fi

\hypersetup{
    linkcolor=eq-fig-tab-color,  
    citecolor=ref-color,  
}
\ifwacvfinal
\else
\fi

\usepackage[sort&compress,numbers]{natbib}
\usepackage[noabbrev,capitalize]{cleveref}
\usepackage[subtle]{savetrees} 

\wacvalgorithmstrack

\begin{document}

\title{Leveraging Local Patch Differences in Multi-Object Scenes\\for Generative Adversarial Attacks}

\author{Abhishek Aich, Shasha Li, Chengyu Song, M. Salman Asif,\\ Srikanth V. Krishnamurthy, Amit K. Roy-Chowdhury\\
University of California, Riverside, USA\\
{\tt\small \{aaich001@, sli057@, csong@cs., sasif@ece., krish@cs., amitrc@ece.\}ucr.edu}
}

\maketitle

\input{sections/0_abstract}

\input{sections/1_introduction}

\input{sections/2_related_works}

\input{sections/3_methodology}

\input{sections/4_experiments}

\input{sections/5_conclusion}


\newpage
\onecolumn
\begin{center}
  \vspace*{2\baselineskip}
  \Large\bf{Supplementary material for ``Leveraging Local Patch Differences in Multi-Object Scenes for Generative Adversarial Attacks''}  
\end{center}
\vspace{6\baselineskip}

\setlength\parindent{0pt}

\input{supp_material/1_add_exps}
\twocolumn
\FloatBarrier
{\small
\bibliographystyle{unsrt}
\bibliography{egbib}
}

\end{document}

%% file: sections/0_abstract.tex
\begin{abstract}
    State-of-the-art generative model-based attacks against image classifiers overwhelmingly focus on single-object (\ie., single dominant object) images. Different from such settings, we tackle a more practical problem of generating adversarial perturbations using multi-object (\ie., multiple dominant objects) images as they are representative of most real-world scenes. Our goal is to design an attack strategy that can learn from such natural scenes by leveraging the local patch differences that occur inherently in such images (\eg. difference between the local patch on the object `\textit{person}' and the object `\textit{bike}' in a traffic scene). Our key idea is to misclassify an adversarial multi-object image by confusing the victim classifier for each local patch in the image. Based on this, we propose a novel generative attack (called Local Patch Difference or LPD-Attack) where a novel contrastive loss function uses the aforesaid local differences in feature space of multi-object scenes to optimize the perturbation generator. Through various experiments across diverse victim convolutional neural networks, we show that our approach outperforms baseline generative attacks with highly transferable perturbations when evaluated under different white-box and black-box settings. 
\end{abstract}

%% file: sections/1_introduction.tex
\section{Introduction}
\input{figures/teaser}
\label{sec:intro}
Understanding and exposing security vulnerabilities of deep neural networks (DNNs) has been an important recent focus of the computer vision research community \cite{jia2019object, zhu2020method, bogdan2017method, duan2020adversarial}. DNNs have been extremely effective in recognition and classification systems like pedestrian recognition \cite{szarvas2005pedestrian, lwowski2017pedestrian, Aich_2021_ICCV} and health-care applications \cite{shkolyar2019augmented, cruz2017accurate}. Images of real-world scenes usually consist of multiple objects. Such scenes are often analyzed by classifiers which predict all the object labels present in such images for downstream tasks such as object annotation \cite{boutell2004learning, zhang2007ml, read2011classifier, chen2018order, allwein2000reducing, MLGCN_CVPR_2019}. Since DNNs are known to be vulnerable to adversarial attacks, it is important to understand the vulnerabilities of such multi-object classifiers. For example, scenes monitored by drones can be attacked by adversaries where all object labels detected are changed for misinterpretation at the user end \cite{varghese2018changenet}. Investigating such scenarios where the multi-object classifiers fail is important in order to design robust and secure real-world systems.

\indent Adversarial attacks can be broadly classified as instance-driven approaches that are image (i.e. instance) specific \cite{goodfellow2014explaining, carlini2017towards, moosavi2016deepfool}) and the distribution-driven or generative model-based approaches (\eg. \gap, \cda, and \tda). Generative model attacks learn to craft perturbations by a generative model via training on a data distribution against a surrogate classifier. Victim classification attributes (\eg. kind of model architecture, data distribution, \etc) are generally unknown by attackers in practical cases. Hence, attackers aim towards creating strong transferable perturbations. Generative attacks provide this distinct advantage over instance-driven attacks for better transferability of perturbations for attacking unseen models \cite{naseer2019cross} as well as better time complexity \cite{xiao2018generating, naseer2019cross, zhang2022beyond, mopuri2018nag, poursaeed2018generative}. Our work focuses on generative attacks for learning to create perturbations using multi-object images and disrupt all labels predicted by victim classifiers. For example in \cref{fig:teaser}, we aim to change the labels associated with the image (\ie `\textit{person}' and `\textit{bike}') to labels whose objects do not exist in the input image with imperceptible perturbations (\eg `\textit{car}', `\textit{dog}'). Existing generative model attacks (see \cref{tab:related-works}) typically attempt to perturb images with a single dominant object in them which are analyzed by single-label classifiers. Using such single-object attacks on multi-object images would require independent object binary segmentation masks to focus on every single object in order to perturb them. This makes these attacks inefficient and impractical as an attacker cannot assume to have object binary masks for every possible distribution on the victim end. 

\indent The focus of this paper is to learn to create perturbations on multi-object images that can disrupt the output of various victim (multi-object or single-object) classifiers for \textit{all} labels of the input image, \textit{without} any need for independent attacks on individual objects. To this end, we propose a novel attack method that utilizes the local difference of patches in the multi-object image. As multi-object images generally contain multiple dominant objects, it is highly likely that the majority of the patches sampled are from different objects. Based on these ``inherent local differences" in multi-object images, we propose a method that utilizes this property to train a perturbation generator.

Our core idea is: if the object is to be misclassified, a patch over the object should also be misclassified (in other words, make them ambiguous to the victim model). To create this misclassification, we exploit the rich local patch differences provided by multi-object images and train a perturbation generator using a novel contrastive learning loss. More specifically, given an image with multiple objects (
\eg `\textit{bike}', and `\textit{person}' in \cref{fig:teaser}), we aim to use the \textit{local} difference of the feature patch on object `\textit{bike's tire}' and the feature patch on object `\textit{person's head}'. Assuming the size of clean and perturbed image are the same (\eg. 224 $\times$ 224), our proposed contrastive strategy \textbf{misaligns} a query patch from feature map of perturbed image (say patch from `\textit{person's head}') with the patch from corresponding or the same location on feature map of a clean image, by simultaneously \textbf{aligning} it with patches from non-corresponding or different locations (say patch from `\textit{bike's tire}' and `\textit{bike's engine}') on feature map of clean image. Our \textbf{intuition} to do so is: we want the feature patch on `\textit{person's head}' in the perturbed image to change to some random features in order to create ambiguity and eventually confuse the victim classifier. 

Unique to multi-object images, this location information is readily available in them due to the spatial arrangement of objects, \textit{without} the need for any kind of labels or segmentation maps. Further, local patches (on average) differ from each other even if they belong to the same object, \eg the shape of the engine of a bike will differ from the shape of the tyre. 

\indent Our approach is fundamentally different from prior single-label image based generative attack approaches \cite{nakka2020indirect, naseer2019cross, poursaeed2018generative} which do not use any kind of aforesaid \textit{local} differences in feature maps of clean and perturbed images. Specifically, we use the approach of contrastive learning where the perturbation generator learns to disassociate corresponding signals of clean and perturbed image features, in contrast to other non-corresponding signals. In our case, these corresponding signals are patches at the same spatial location in clean and perturbed image features, while non-corresponding signals are patches at different spatial locations in the clean image features. The contrastive learning approach has been extensively used in unsupervised learning \cite{park2020contrastive, he2020momentum, wu2018unsupervised, andonian2021contrastive} for various image downstream tasks. We demonstrate its benefits in optimizing perturbation generating models for highly potent adversarial attacks. We refer to our attack approach as \textbf{L}ocal \textbf{P}atch \textbf{D}ifference attack or \game (see Figure \ref{fig:main-fig}). \game uses our novel local-patch contrasting approach and learns to create strong imperceptible perturbations on multi-object images. 
\newline
\indent To validate our approach, we evaluate \game's generated perturbations in different challenging scenarios. For example, if a perturbation generator is trained on \voc \cite{everingham2010pascal} dataset with a Res152 \cite{he2016deep} \voc pre-trained multi-object classifier as a surrogate, then from the attacker's perspective, we show that \game crafts highly transferable perturbations under following settings (in order of least realistic to most) 
\begin{enumerate}[topsep=0.5em, align=left, leftmargin=0em, itemindent=!, label = \underline{$\bullet$ \textit{Setting} \arabic*.}]
\setlength\itemsep{0em}
\item \textit{white-box}: victim classifier is \textit{seen}, victim data dataset is \textit{seen}, victim task is \textit{seen} (\eg Res152 multi-object classifier, \voc dataset, multi-object classification task)
\item \textit{black-box}: victim classifier is \textit{unseen}, victim data dataset is \textit{seen}, victim task is \textit{seen} (\eg VGG19 multi-object classifier, \voc dataset, multi-object classification task)
\item \textit{strict black-box}: victim classifier is \textit{unseen}, victim data dataset is \textit{unseen}, victim task is \textit{seen} (\eg VGG19 multi-object classifier, \coco \cite{lin2014microsoft} dataset, multi-object classification task)
\item \textit{extreme black-box}: victim classifier is \textit{unseen}, victim dataset is \textit{unseen}, victim task is \textit{unseen} (\eg VGG16 single-label classifier, ImageNet \cite{krizhevsky2012imagenet} dataset, single-label classification task)
\end{enumerate}
`\textit{Setting} 4' is especially useful to test the strength of crafted perturbations by different attacks because it presents real-world use case for attackers where all victim attributes like classifier architecture, data distribution and task is unseen.
To summarize, we make the following \textit{contributions}:
\begin{enumerate}[topsep=0.5em, leftmargin=0em, align=left, itemindent=!, label = \arabic*.]
\setlength\itemsep{0em}
    \item \textbf{New practical problem.} \tcomment{We tackle a new problem of learning to craft perturbations for multi-object data distributions, the situation in most real-life scenes, using generative model-based attacks to disrupt decisions. To the best of our knowledge, this is the first work that considers to create \textbf{generative attacks} using multi-object images.}%
    \item \textbf{Novel attack framework.} \tcomment{To this end, we propose a novel generative model-based attack approach namely \game, where the perturbation generator is trained using a contrastive loss that uses rich local patch differences of multi-object image features.} %
    \item \textbf{Extensive experiments.} Through extensive experiments on two multi-object benchmarks, we show that \game has overall better attack transferability and outperforms its baselines under aforementioned settings (see \cref{tab:summ_voc} and \cref{tab:summ_coco}).
\end{enumerate}

%% file: figures/teaser.tex
\begin{figure}[ht]
  \centering
  \includegraphics[width=\columnwidth]{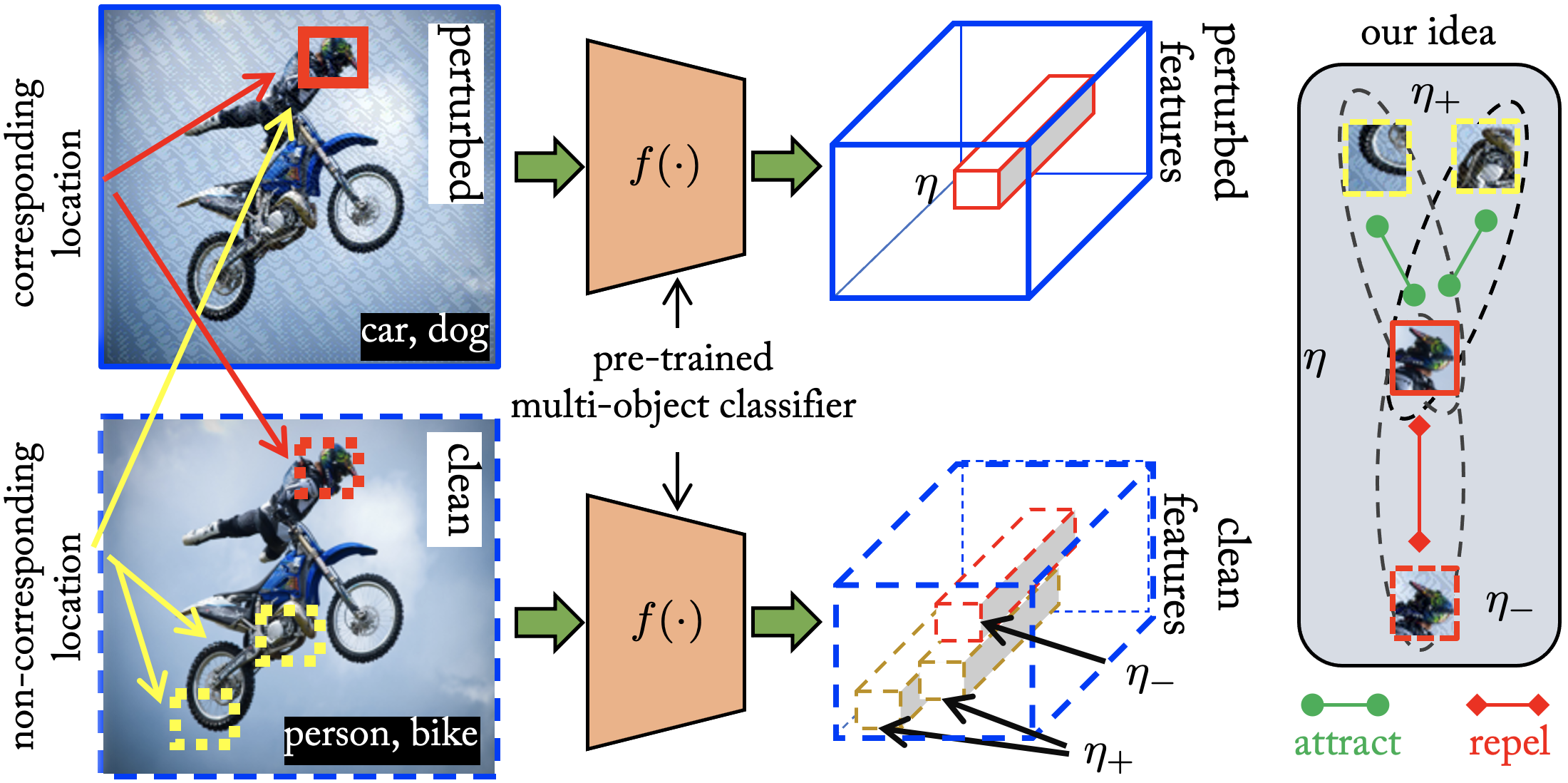}
  \caption{\textit{Proposed attack \game:} We aim to create perturbations using multi-object images. To do this, our proposed attack \game leverages the rich \textit{local} differences between the patches of features extracted from multi-object images. \eg, the local feature patch of `\textit{person's head}' will be different from local feature patch of `\textit{bike's tire}' or `\textit{bike's engine}'. \tcomment{\game leverages these differences to misalign (repel) a query patch ($\eta$) from perturbed image feature with the corresponding patch ($\eta_-$) from clean image feature, while aligning (attract) with non-corresponding patches of different locations ($\eta_+$)}.} 
  \label{fig:teaser}
\end{figure}

%% file: sections/2_related_works.tex
\section{Related Work}
\vspace{0.8em}
\label{sec:related-work}
\input{figures/main_fig}
\paragraph{\textbf{Adversarial attacks on image classifiers}.} Most existing state-of-the-art adversarial attack works \cite{fawzi2018analysis, goodfellow2014explaining, kurakin2016adversarial, nguyen2015deep, naseer2019cross, poursaeed2018generative,liu2019perceptual, han2019once, nakka2020indirect, carlini2017towards, xiao2018spatially, zhang2022beyond, xiao2018generating, li2021adversarial} have been designed to attack single-object classifiers. Among these attacks, instance (or image)-driven perturbations \cite{fawzi2018analysis, goodfellow2014explaining, kurakin2016adversarial, nguyen2015deep, fan2020sparse} have been extensively explored, both to showcase the various shortcomings of single-object classifiers  \cite{szegedy2013intriguing}. Instance-driven attacks are characterized by their method of computation of perturbations only on corresponding clean images. This results in perturbations being computed for each image individually, without using knowledge from other images \cite{naseer2019cross}. The current literature on instance-driven approaches broadly consists of methods that use gradient ascent on the images \cite{goodfellow2014explaining, dong2018boosting, moosavi2016deepfool, fan2020sparse} or the those that generate adversarial examples using optimization-based methods \cite{carlini2017towards, xiao2018spatially} for attacking single-object classifiers. Attacks on multi-object classifiers using instance-driven approaches have been proposed in \cite{song2018multi, zhou2020generating, hu2021tkml}. \cite{zhou2020generating} proposed a method to create multi-object adversarial examples by optimizing for a linear programming problem. \cite{song2018multi} proposed a method to exploit label-ranking relationships based framework to attack multi-object ranking algorithms. More recently, \cite{hu2021tkml} presented a method to disrupt the top-$k$ labels of multi-object classifiers. Although effective for perturbing single images, instance-driven approaches are inefficient when it comes to attacking a large dataset of images, as the perturbations will have to be generated by iterating over these images individually multiple times \cite{naseer2019cross, zhang2022beyond}. Different from \cite{song2018multi, zhou2020generating, hu2021tkml}, \game falls under the category of \textit{generative model-based} adversarial attacks (which we discuss next) that are distribution-driven approaches. Such approaches train a generative network over a large number of images to create perturbations. Once the model is trained, it can be used to perturb multiple images simultaneously.
\paragraph{\textbf{Generative model-based adversarial attacks.}} To address the shortcomings of instance-driven approaches, generative model-based or distribution-driven attack approaches \cite{naseer2019cross, poursaeed2018generative, salzmann2021learning, xiao2018generating, lu2021discriminator, zhang2022beyond, mopuri2018nag} have been explored recently for learning perturbations on single-object images. For example, \gap presents a distribution-driven attack that trains a generative model for creating adversarial examples by utilizing the cross-entropy loss. Recently, \cda proposed a generative network that is trained using a relativistic cross-entropy loss function. Both \gap and \cda rely on the final classification layer of the surrogate model to train the perturbation generator which has been shown to have inferior transferability of perturbations to unknown models. Different from these, \cite{salzmann2021learning} presented an attack methodology to enhance the transferability of perturbations using feature separation loss functions (\eg mean square error loss). However, their attack requires a manual selection of a specific mid-layer for every model against which the generator is to be trained. In contrast to these aforementioned works, \game is designed to learn to craft imperceptible adversarial perturbations using \textit{multi-object} images. Rather than focusing on the feature map globally, we take a more fine-grained approach of (feature map) patch contrasting via a novel contrastive loss. More specifically, \game uses the local feature differences at multiple mid-level layers and uses an InfoNCE loss \cite{van2018representation} based framework to create highly effectual perturbations. We summarize the differences of \game with the aforementioned generative attack methods in \cref{tab:related-works}. 
\input{tables/related_works}

%% file: figures/main_fig.tex
\begin{figure*}[t]
  \centering
  \includegraphics[width=\textwidth]{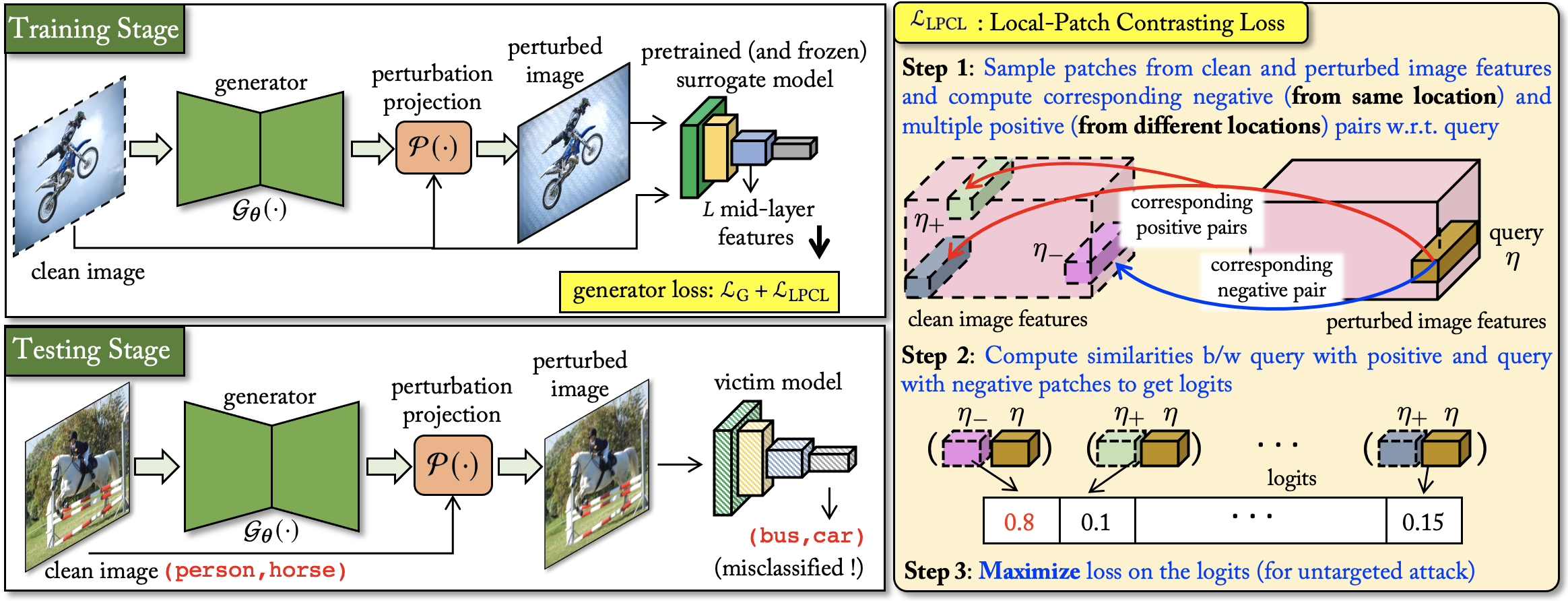}
  \caption{\textit{Framework overview.} Our proposed \game framework (\textit{top}) aims to learn from multi-object images using a contrastive learning mechanism ($\ccl$) to maximize the difference of corresponding patches of same locations while minimizing the difference between non-corresponding patches of distinct locations, from features extracted from clean and perturbed images. This results in highly effective and transferable perturbations for input clean images during inference (\textit{bottom-left}).} 
  \label{fig:main-fig}
  \vspace{-0.5\baselineskip}
\end{figure*}

%% file: tables/related_works.tex
\begin{table}[t]
\setlength{\tabcolsep}{0.5pt}
\renewcommand{\aboverulesep}{0pt}
\renewcommand{\belowrulesep}{0pt}
\caption{\textit{Characteristic comparison.} Better than prior generative attacks \cite{poursaeed2018generative, naseer2019cross, salzmann2021learning}, \game is a generative attack method designed for ``multi-object" images. Here, CE$(\cdot)$: Cross-Entropy loss, MSE$(\cdot)$: Mean-Square Error loss, $\bm{f}$: surrogate classifier used for training perturbation generator $\pertG$ (weights $\wG$). $\x$: clean image, $\pertx$: perturbed image, and $\pert$: perturbation. $\ell$: output from specific pre-defined layer. $t$: misclassification label depending on type of attack (targeted or untargeted). Proposed loss $(\mse + \ccl)$ is detailed in \cref{sec:method}.}
\centering
\resizebox{\columnwidth}{!}{
\small
\begin{tabularx}{\columnwidth}{c*{2}{Y}c*{2}{Y}}
\toprule
\rowcolor{red!10}
 & & \multicolumn{2}{c}{\textbf{Classifier Attack Strategy}} \\ 
\Xcline{3-4}{0.05em} 
\rowcolor{red!10}
 \multirow{-2}{*}{\dual{\textbf{DD}/\textbf{Attacks}}} & \multirow{-2}{*}{\textbf{Venue}} & \textbf{image type}?  & $\pertG$ \textbf{loss} \\ \midrule
 \gap & CVPR2018 & single-object & CE($\bm{f}$($\pertx$), $t$) \\
 \cda & NeurIPS2019 & single-object & CE($\bm{f}$($\pertx$) - $\bm{f}(\x)$, $t$) \\
 \tda & NeurIPS2021 & single-object & MSE($\bm{f}_{\ell}$($\pertx$), $\bm{f}_{\ell}(\x))$ \\
 \midrule
 \game & Ours & multi-object & $\mse + \ccl$\\
 \bottomrule
\end{tabularx}}
\label{tab:related-works}
\end{table}

%% file: sections/3_methodology.tex
\section{Proposed Attack Methodology}
\label{sec:method}
Here, we explain our proposed generative adversarial attack \game that learns from multi-object images. It includes training the perturbation generator with a novel local patch contrasting learning loss that uses local regions of features extracted from clean and perturbed images. We start with the notations and defining the problem statement. 

\subsection{Problem Formulation}
\vspace{0.8em}
\paragraph{\textbf{Notations}.} Let $C$ be the total number of classes and $N$ be the number of training samples in a dataset $\mathcal{T}$. We define $\mathcal{T} = \{(\x^{(1)}, \bm{y}^{(1)}), \cdots, (\x^{(N)}, \bm{y}^{(N)})\}$ where $\x^{(i)}\in\mathbb{R}^{H\times W\times C}$ and  $\bm{y}^{(i)} = [y_1^{(i)}, \cdots, y_C^{(i)}]\in \mathcal{Y} \subseteq \{0, 1\}^C$ are the $i$th image (with height $H$, width $W$, and channels $Z$) and ground-truth label vector, respectively. For an example data point $\x^{(i)}$ and class $c$, $y_c^{(i)}=1$ (or $= 0$) indicates the presence (or absence) of an object from class $c$ in $\x^{(i)}$. We define a surrogate multi-object classifier trained on $\mathcal{T}$ as $\f$, which is utilized to train a perturbation generator $\pertG$ (parameterized by weight $\wG$). In further discussions and \cref{fig:main-fig}, we drop the superscript $i$ for ease of exposition.
\paragraph{\textbf{Problem Statement}.} Given a clean multi-object image $\x$ from data-distribution $\mathcal{T}$ containing multiple dominant objects and the victim classifier $\g$, we aim to flip \emph{all} labels of $\x$ with an allowable perturbation budget $\epsilon$ defined by an $\ell_\infty$ norm. Specifically, the problem objective is to craft a perturbation $\pert$ such that the prediction of $\g$ for all labels $\bm{y}$ associated with $\x$ is changed. Mathematically, this can be represented as $\bm{y} \neq \widehat{\bm{y}}$ where, $\bm{y} = \bm{g}\big{(}\x\big{)}$ and $ \widehat{\bm{y}} = \bm{g}\big{(}\x+\pert\big{)}$ with $\Vert\pert\Vert_\infty \leq \epsilon$.
\subsection{Proposed Approach: \game}
Our proposed framework is presented in \cref{fig:main-fig}. It contains a perturbation generator $\pertG$ that is trained to craft imperceptible perturbations $\pert$ on $\x$. $\pertG$ is trained against a surrogate pre-trained multi-object classifier $\f$. More precisely, $\f$ acts as a discriminator against which generator $\pertG$ is trained ($\f$ remains fixed or frozen). During training, $\pertG$ takes $\x$ as input and generates an unbounded perturbed image $\mathcal{G}(\x) = \widehat{\pertx}$. This unbounded perturbed image $\widehat{\pertx}$ is clipped to be within an pre-defined perturbation budget $\epsilon$ on $\x$ under the $\ell_\infty$ norm using the projection operator $\mathcal{P}(\cdot)$. The perturbed image is then estimated as $\pertx = \mathcal{P}(\widehat{\pertx})$. To compute the generator loss, $\pertx$ is sent to the discriminator, $\f$, to be misclassified. At multiple $L$ mid-layers from $\f$, we compute the features of clean image $\big{[}\bm{f}_k(\x)\big{]}_{k=1}^L$ and features of perturbed image $\big{[}\bm{f}_k(\pertx)\big{]}_{k=1}^L$, where $\bm{f}_k(\x), \bm{f}_k(\pertx)\in \mathbb{R}^{h_k\times w_k \times c_k}$. Here, $h_k\times w_k$ denote the spatial size of $i$th layer feature map with $c_k$ channels. The effectiveness of using mid-level features to craft powerful perturbations have been extensively studied in \cite{nakka2020indirect, zhang2022beyond, zhou2018transferable, DBLP:conf/nips/InkawhichLWICC20, inkawhich2019feature, li2021transrpn}. Therefore, we leverage these mid-level features of $\f$ and define our generative model loss via two functions. The \textit{first} loss function is a global loss $\mse$ that compares extracted features directly as follows:
\begin{align}
    \mse = \dfrac{1}{L}\sum_{k=1}^L\text{dist}\big{(} \bm{f}_k(\x),\bm{f}_k(\pertx) \big{)}
\end{align}
Here, dist($\cdot$) can be any distance measuring function, \eg mean square error function, etc. The \textit{second} loss function is a novel objective, namely, Local-Patch Contrasting Loss (LPCL) which compares the extracted features $\big{[}\bm{f}_k(\x)\big{]}_{k=1}^L$ and $\big{[}\bm{f}_k(\pertx)\big{]}_{k=1}^L$ at a local or patch level. Better than prior generative attacks which only compare perturbed and clean images globally, our proposed LPCL loss leverages the local difference of patches from multiple objects in the input image to disrupt the victim classifier's decisions. We expand on the details of LPCL next.  

\input{tables/game_training}
\subsection{Contrasting Patches of Multi-Object Images}
\vspace{0.8em}
\label{sec:explain-context}
\paragraph{\textbf{Motivation}.} We make the observation that due to existence of multiple objects in a multi-object image $\x$, we can utilize the local feature patches from $\bm{f}_k(\x)$ (and $\bm{f}_k(\pertx)$). The local patches of input clean image belong to individual dominant objects and thus, prompt the multi-object classifier to output their respective associated labels. Therefore, for each object in a perturbed image to be misclassified, each patch within its feature map should look different to the classifier than the same location corresponding patch in the feature map of a clean image. To create this difference, we use the feature maps from different location non-corresponding patches to create ambiguity for the victim classifier to prompt incorrect decisions on the overall perturbed image. This patch location-wise contrasting of the clean and perturbed image features at the local level allows for stronger supervision for training the perturbation generator $\pertG$. 

\paragraph{\textbf{Proposed contrasting loss} ($\ccl$).} 
To misclassify the perturbed image $\pertx$, we need to maximize the difference between its features and that of the clean image $\x$. We propose to achieve this by misaligning corresponding clean-perturbed image feature patches at a specific location to maximize the difference at a local level. This misalignment is enabled by utilizing the other patches from the clean image features at non-corresponding locations. We start with computing the features of clean and perturbed image from surrogate model $\f$ as $\big{[}\bm{f}_k(\x)\in \mathbb{R}^{h_k\times w_k \times c_k}\big{]}_{k=1}^L$ and $\big{[}\bm{f}_k(\pertx)\in \mathbb{R}^{h_k\times w_k \times c_k}\big{]}_{k=1}^L$, respectively. We convert these feature maps to tensors $\bm{D}_k$ and $\widehat{\bm{D}}_k$, respectively, of size $v_k\times c_k$ (where, $v_k=h_kw_k$). Next, we chose a query vector $\bm{\eta}^q_k\in\mathbb{R}^{c_k}$ from a $q$th spatial location of $\widehat{\bm{D}}_k$ and choose the corresponding spatial location vector $\bm{\eta}^{-}_{k}$ from $\bm{D}_k$, which we call $\widehat{\bm{\eta}}_k$'s negative. Then, from $R$ other (or different) locations of $\bm{D}_k$, we choose a collection of positives denoted by $\bm{\eta}^+_k\in\mathbb{R}^{R\times c_k}$. The $\ccl$ loss is now defined as an $(R+1)$-way classification objective, with logits representing the similarity between query $\bm{\eta}^q_k$ and set $[\bm{\eta}^-_k,\bm{\eta}^+_{k1},\bm{\eta}^+_{k2}, \cdots, \bm{\eta}^+_{kR}]$, as follows.
\begin{equation}
    \Scale[0.7]{
    \ccl = -\dfrac{1}{L}\sum_{k=1}^L\log\bigg{(}\dfrac{\exp\big{(}sim(\bm{\eta}^q_k,\bm{\eta}^-_k)\big{)}}{\exp\big{(}sim(\bm{\eta}^q_k,\bm{\eta}^-_k)\big{)} + \sum_{r=1}^R\exp\big{(}sim(\bm{\eta}^q_k,\bm{\eta}^+_{kr})\big{)}}\bigg{)}
    }
\end{equation}
where $sim(\bm{\eta}_a,\bm{\eta}_b) = \nicefrac{\bm{\eta}_a^\top\bm{\eta}_b}{\tau}$ returns the similarity between two vectors, $\top$ represents the transpose operation, and $\tau$ is a scaling parameter. We set $\tau = 0.07$ following \cite{park2020contrastive}. This loss envisions our idea that, if a feature patch on the perturbed image is to be disrupted, it should obtain a low similarity score with the corresponding (same location) ``negative" feature patch of the clean image, and high similarity score with ``positive" patches from non-corresponding locations. Note that ``patch" does not correspond to ``object" and it is possible that (\textit{1}) group of patches can belong to one object and (\textit{2}) one patch can contain parts of multiple objects. The only requirement for the $R$ positive patches used in $\ccl$ to operate properly is: these $R$ positive patches should contain feature values that are different from the values in query feature patch $\bm{\eta}^q_k$. This requirement is easily fulfilled when we sample them from non-overlapping \wrt to each other and from different locations \wrt to $\bm{\eta}^q_k$.  
\subsection{Final Objective}
Our final learning objective includes a loss function to train the generator over $\pertx$ both \textit{globally} with a $\mse$ objective and \textit{locally} using our proposed contrasting loss $\ccl$. This loss is computed over multiple $L$ mid-level layers of $\f$ as $\mathcal{L} = \mse + \ccl$. Note that we maximize $\mathcal{L}$ for an untargeted attack with $\mse$ set as the mean square error loss. For targeted attack, we minimize $\mathcal{L}$ with $\mse$ set as binary cross-entropy loss to classify the perturbed image to the target label. The whole training procedure is summarized in \cref{alg:train}. During testing, we simply input the test image to the trained generator to create a perturbed image with the aim to fool the victim classifier for all the associated labels.

%% file: tables/game_training.tex
\begin{algorithm}[t]
    \small
	\DontPrintSemicolon
	\caption{\game Training Algorithm}
    \label{alg:train}
    \KwInput{clean images $\x$ from distribution $\mathcal{T}$, perturbation $\ell_\infty$ bound $\epsilon$, surrogate classifier $\f$}
    \KwInput{learning rate $\alpha$}
    \KwOutput{perturbation generator $\pertG$'s weights $\wG$}
    \smallskip
    
    \tcc{Large-Scale training of $\pertG$}
    Randomly initialize $\wG$\\
    Load and freeze multi-object classifier $\f$ trained on $\mathcal{T}$ \\ 
	\While{not done}
	{
	    \tcc{Obtain clean image features}
		Input $\x$ to $\f$ and get $L$ mid-layer features $\big{[}\bm{f}_k(\x)\big{]}_{k=1}^L$\\
		\tcc{Obtain perturbed image features}
		Create unbounded perturbed image $\mathcal{G}(\x)$ \\
		Project it within bound $\epsilon$ using $\mathcal{P}(\cdot)$ to obtain $\pertx$ \\
		Input $\pertx$ to $\f$, get $L$ mid-layer features $\big{[}\bm{f}_k(\pertx)\big{]}_{k=1}^L$\\
		\tcc{Compute loss}
		Compute $\mathcal{L}=\mse+\ccl$ \\
		\tcc{Update $\pertG$'s weights}
		Update $\wG$ with respect to $\mathcal{L}$ using Adam\\
		$\wG\leftarrow\wG - \alpha \nabla_{\wG} \mathcal{L}(\wG)$
	}
\end{algorithm}

%% file: sections/4_experiments.tex
\section{Experiments and Results}
\label{sec:exp}
Here, we discuss the strength of \game under diverse attack \textit{Settings 1-4} (as described in \cref{sec:intro}) presented in \cref{tab:summ_voc} and \cref{tab:summ_coco}. Furthermore, we analyze the strength of \game on most realistic attack setting in \cref{tab:voc_sl} and \cref{tab:coco_sl}, as well as other easier variations in \cref{tab:main_voc}. We also perform an ablation analysis of \game in \cref{fig:abl_analysis} and show some examples of perturbed images and attention shift in \cref{fig:qual_result} to validate our method. Unless otherwise stated, perturbation budget is set to $\ell_\infty\leq10$. We provide details of implementation, baselines (\gap, \cda, \tda), and additional experiments in the Supplementary Material.
\begin{description}[leftmargin=0em, itemsep=0.25ex, labelwidth=*, itemindent=!]

\item[Training Datasets.] We employ widely-used and publicly available PASCAL-VOC \cite{everingham2010pascal} and MS-COCO \cite{lin2014microsoft} datasets. For \voc, we use \textit{trainval} from `VOC2007' and `VOC2012' as our training dataset and the evaluations are carried out on `VOC2007\_test' dataset. For \coco, we use \textit{train2017} as our training dataset and \textit{val2017} for evaluations.

\item[Inference Metrics.] We evaluate the attacks on multi-object classifiers using accuracy on test set defined for multi-object classification in \cite{godbole2004discriminative, sorower2010literature}. For attacks on single-object classifier in \cref{tab:voc_sl} and \cref{tab:coco_sl}, we use top-1 accuracy on test set. For all untargeted attacks, a lower score indicates better attack. In case of targeted attack, a higher score indicates better attack result. Best results are in \tred{red}, second best are in \tblue{blue}. Accuracy on clean images are provided in \tgray{gray} for reference.
\item[Victim Models and Attack Settings.] \tcomment{To attack the victim models, we first train all the perturbation model $\pertG$ for baselines and \game on \voc and \coco on their respective train set against surrogate multi-classifier model $\f$. We chose $\f$ to be (\voc or \coco) pre-trained multi-object classifiers Res152 \cite{he2016deep}, Dense169 \cite{huang2017densely}, and VGG19 \cite{simonyan2014very}. As discussed in \cref{sec:intro}, we then evaluate the trained $\pertG$ under four following settings. Firstly for \textit{Setting} 1 (white-box), we attack the surrogate multi-classifier model $\f$ on test set of \textit{same} multi-object distribution used during training. Secondly for \textit{Setting} 2 ({black-box}), we attack \textit{other} multi-object classifiers different from the surrogate model also on test set of \textit{same} multi-object distribution used during training. Thirdly for \textit{Setting} 3 ({strict black-box}), we attack multi-object classifiers on test set of \textit{different} multi-object distribution \textit{other than} used during training. Finally, following \cite{zhang2022beyond} for \textit{Setting} 4 ({extreme black-box}), we attack various single-object classifiers for CIFAR10 \cite{krizhevsky2009learning}, CIFAR100 \cite{krizhevsky2009learning}, STL-10 \cite{coates2011analysis}, and SVHN \cite{netzer2011reading} (coarse-grained tasks), CUB-200-2011 \cite{wah2011caltech}, Stanford Cars \cite{krause2013collecting}, and FGVC Aircrafts \cite{maji2013fine} (fine-grained tasks), and ImageNet \cite{deng2009imagenet} models on their respective test sets}. The pre-trained victim models of coarse-grained tasks are available in \cite{coarse2022task}, for fine-grained tasks (Res50 \cite{he2016deep} and SENet154 \cite{hu2018squeeze}) in \cite{fine2022task} and ImageNet task in \cite{imagenet2022task}. Briefly, the coarse-grained single-object classification task is to distinguish labels like `cats \textit{vs} dogs', whereas the fine-grained single-object classification task is to distinguish difficult labels like species of cats (\eg `tiger \textit{vs} panther'). Analyzing attacks on such diverse tasks after learning perturbations from multi-object images will show the transferability of perturbations which is important for real-world attacks.
\end{description}
\input{tables/summary_table_voc}
\input{tables/summary_table_coco}
\subsection{Quantitative Results}
We evaluated \game against baselines under four different attack scenarios. We summarize them in \cref{tab:summ_voc} and \cref{tab:summ_coco}, and discuss them below.

\begin{description}[leftmargin=0em, labelwidth=*, itemsep=0ex, itemindent=!] 
\item[Observation 1.] \textit{The proposed method \game has the overall best performance.} We outperform the prior best SOTA method \tda in 10 out of 12 cases, demonstrating the efficacy of our proposed method. For example in \voc, we outperform TDA by a margin of ~10\% (ours: 55.88\%, TDA: 65.08\%), and in \coco, by a margin of ~3.5\% (ours: 46.73\%, TDA: 51.00\%). Furthermore, TDA carries an expensive computational overhead (discussed by the authors themselves in Section 4.6 under ``Limitations''): the attacker needs to incur high time complexity (by training the generator separately for each possible mid-layer) to search for the most effective mid-layer of the surrogate model in order to optimize the generator. Through our results, especially on the ImageNet dataset, we show that TDA’s manually selected specific layer is highly sensitive to the training data distribution as the results on ImageNet degrade drastically if the generator is trained on datasets different from ImageNet (in this case, \voc, \coco). In contrast, since we select a group of layers, we do not need this laborious time and resource-consuming analysis.
\item[Observation 2.] \textit{SOTA tends to comparatively overfit more to the attacker’s training data distribution than the proposed method}. The aforementioned four attack scenarios (after the generator is trained on \voc and \coco) show that: as the victim data distribution starts varying (\eg. ImageNet, STL-10, FGVC Aircraft classification), there is a huge performance drop in the prior attacks due to weaker transferability of perturbations. For example, TDA shows a comparable performance when the victim distribution is similar to the attacker’s training distribution (see \cref{tab:main_voc}) but shows surprisingly low attack results (20\% difference) when the victim distribution changes to single-object classifications tasks like ImageNet, STL-10, FGVC Aircraft (see \cref{tab:voc_sl}). This clearly demonstrates that prior works tend to overfit to the attacker's training distribution and perform poorly when there is no overlap in the victim’s data distribution and type of classification task. On the other hand, our proposed method \game alleviates this issue and shows better transferability of perturbations. We attribute the better performance of our method, in better alleviating the overfitting issue than SOTA, to the unique strategy of comparing local feature patches rather than just global differences. 
\item[Observation 3.] \textit{As attack scenarios become more difficult and realistic, the proposed method's performance is much better than the SOTA baselines}. White-box attacks (\textit{Setting} 1) are easy and least realistic attacks, whereas extreme black-box attacks (\textit{Setting} 4) are the most difficult but most realistic (the attacker has no knowledge of the victim model or task) attack settings. We observe that as the difficulty level of attack increases, the performance of TDA crafted perturbations show increasingly poor performance than the proposed method \game. For example, though \game and TDA show comparable performance in the white-box attacks, it outperforms TDA by a huge margin of ~18\% in extreme black-box attacks (see \cref{tab:coco_sl} and \cref{tab:voc_sl}). This implies existing attacks perform poorly in real-world use cases, whereas \game poses a greater threat to the victim model than prior SOTA attacks.
\input{tables/cd_method_comparison_voc}
\input{tables/cd_method_comparison_coco}
\input{tables/wb_bb_method_comparison_voc}
\item[Targeted attacks.] We performed a white-box targeted attack on Dense169 with the target label set to `person' (\ie all perturbed images should output the label `person'). We observed that \gap and \cda result in an accuracy of 34.58\% and 34.86\% whereas \game resulted in 35.00\% attack performance (perturbation bound $\ell_\infty\leq16$).
\end{description}
\input{figures/ablation/ablation_analysis}
\subsection{Ablation Study} 
We perform an ablation analysis of \game with respect to loss objectives in \cref{fig:loss_abl}, impact of number of patches $R$ in \cref{fig:patches}, and impact of number of layers $L$ in \cref{fig:layers} utilized from the surrogate model $\f$ to train $\pertG$. From \cref{fig:loss_abl}, we observe the impact of components of our loss objective when $\pertG$ was trained against Res152 on \voc both for white-box (test against \voc) and strict black-box (test against \coco). 
It can be observed that the perturbations are most effective when both the global loss $\mse$ and local loss $\ccl$ are utilized. Next from \cref{fig:patches}, we observe that the best performance is observed with $R=256$ patches (note that we use $R=128$ for a slightly better training time-accuracy trade-off). Finally, we analyze the impact of using multiple mid-level features from $\f$ and observe that $L=4$ results in best attacks as it allows the use of diverse features to learn the perturbations. This also shows that we do not need to manually choose a specific layer for better attacks as in the case of \tda, and an average choice of a group of layers creates effective attacks.

\input{figures/qual_result}
\subsection{Qualitative Results} 
We visualize some examples of perturbed images and shift in attention (using CAM \cite{zhou2016learning}) for misclassified images from clean images in \voc and \coco in \cref{fig:qual_result} for Res152 multi-object classifier. It can be observed that \game changes the focus of the victim classifier to irrelevant regions leading to highly successful attacks.

%% file: tables/summary_table_voc.tex
\begin{table}[!t]
\centering
\caption{\textit{Average Results when $\pertG$ trained with \voc}. We summarize the attack capability of prior generative attack works under various victim scenarios with training data as \voc. Results are averaged over three surrogate classifiers for all methods.}
\label{tab:summ_voc}
\resizebox{\columnwidth}{!}{%
\begin{tabular}{cccc}
\hline
\rowcolor{red!10}
\textbf{Attack} &
  \multicolumn{1}{c}{\textbf{Victim Details}} &
  \textbf{Method} &
  \textbf{Mean result} \\
\hline
 &
   &
  \gap &
  55.22 \\
 &
   &
  \cda &
  54.79 \\
 &
   &
  \tda &
  53.73 \\
\hhline{~~*{2}{-}}
\multirow{-4}{*}{\rot{\dual{\textit{Setting} 1/(easy)}}} &
  \multirow{-4}{*}{\dual{\voc/ (victim model = surrogate model)}} &
  \cellcolor[HTML]{FFF2CC}\game &
  \cellcolor[HTML]{FFF2CC}\tred{52.69} 
   \\
 \hline
 &
   &
  \gap &
  56.24 \\
 &
   &
  \cda &
  55.86 \\
 &
   &
  \tda &
  55.32 \\
  \hhline{~~*{2}{-}}
\multirow{-4}{*}{\rot{\textit{Setting} 2}} &
  \multirow{-4}{*}{\dual{\voc / (victim model $\neq$ surrogate model)}} &
  \cellcolor[HTML]{FFF2CC}\game &
  \cellcolor[HTML]{FFF2CC}\tred{54.37} \\
  \hline
 &
   &
  \gap &
  40.86 \\
 &
   &
  \cda &
  40.51 \\
 &
   &
  \tda &
  39.79 \\
  \hhline{~~*{2}{-}}
\multirow{-4}{*}{\rot{\textit{Setting} 3}} &
  \multirow{-4}{*}{\coco} &
  \cellcolor[HTML]{FFF2CC}\game &
  \cellcolor[HTML]{FFF2CC}\tred{38.69} \\
  \hline
 &
   &
  \gap &
  83.96 \\
 &
   &
  \cda &
  82.72 \\
 &
   &
  \tda &
  83.13 \\
 \hhline{~~*{2}{-}}
 &
  \multirow{-4}{*}{\dual{CIFAR(10,100), STL-10, SVHN/(Coarse-Grained tasks)}} &
  \cellcolor[HTML]{FFF2CC}\game &
  \cellcolor[HTML]{FFF2CC}\tred{70.72} \\
  \hhline{~*{3}{-}}
 &
   &
  \gap &
  90.50 \\
 &
   &
  \cda &
  \cellcolor[HTML]{FFFFFF}{\color[HTML]{333333} 90.21} \\
 &
   &
  \tda &
  88.61 \\
  \hhline{~~*{2}{-}}
 &
\multirow{-4}{*}{\dual{\dual{CUB-200, Stanford Cars,/ FGVC Aircraft}/(Fine-Grained tasks)}} &
  \cellcolor[HTML]{FFF2CC}\game &
  \cellcolor[HTML]{FFF2CC}\tred{73.72} \\
  \hhline{~*{3}{-}}
 &
   &
  \gap &
  73.05 \\
 &
   &
  \cda &
  72.33 \\
 &
   &
  \tda &
  69.91 \\
  \hhline{~~*{2}{-}}
\multirow{-12}{*}{\rot{\dual{\textit{Setting} 4/(difficult)}}} &
  \multirow{-4}{*}{ImageNet} &
  \cellcolor[HTML]{FFF2CC}\game &
  \cellcolor[HTML]{FFF2CC}\tred{45.12}\\
\hline

\end{tabular}%
}
\end{table}

%% file: tables/summary_table_coco.tex
\begin{table}[!t]
\centering
\caption{\textit{Average Results when $\pertG$ trained with \coco} We summarize the attack capability of prior generative attack works under various victim scenarios with training data as \coco. Results are averaged over three surrogate classifiers for all methods.}
\label{tab:summ_coco}
\resizebox{\columnwidth}{!}{%
\begin{tabular}{cccc}
\hline
\rowcolor{red!10}
\textbf{Attack} &
  \multicolumn{1}{c}{\textbf{Victim Details}} &
  \textbf{Method} &
  \textbf{Mean result} \\
\hline
 &
   &
  \gap &
  41.09 \\
 &
   &
  \cda &
  39.96 \\
 &
   &
  \tda &
  \tred{34.31} \\
\hhline{~~*{2}{-}}
\multirow{-4}{*}{\rot{\dual{\textit{Setting} 1/(easy)}}} &
  \multirow{-4}{*}{\dual{\coco/ (victim model = surrogate model)}} &
  \cellcolor[HTML]{FFF2CC}\game &
  \cellcolor[HTML]{FFF2CC}34.91
   \\
 \hline
 &
   &
  \gap &
  41.05 \\
 &
   &
  \cda &
  41.17 \\
 &
   &
  \tda &
  37.08 \\
  \hhline{~~*{2}{-}}
\multirow{-4}{*}{\rot{\textit{Setting} 2}} &
  \multirow{-4}{*}{\dual{\coco / (victim model $\neq$ surrogate model)}} &
  \cellcolor[HTML]{FFF2CC}\game &
  \cellcolor[HTML]{FFF2CC}\tred{36.97} \\
  \hline
 &
   &
  \gap &
  56.03 \\
 &
   &
  \cda &
  55.63 \\
 &
   &
  \tda &
  \tred{51.84} \\
  \hhline{~~*{2}{-}}
\multirow{-4}{*}{\rot{\textit{Setting} 3}} &
  \multirow{-4}{*}{\voc} &
  \cellcolor[HTML]{FFF2CC}\game &
  \cellcolor[HTML]{FFF2CC}52.06 \\
  \hline
 &
   &
  \gap &
  84.07 \\
 &
   &
  \cda &
  81.52 \\
 &
   &
  \tda &
  70.40 \\
 \hhline{~~*{2}{-}}
 &
  \multirow{-4}{*}{\dual{CIFAR(10,100), STL-10, SVHN/(Coarse-Grained tasks)}}  &
  \cellcolor[HTML]{FFF2CC}\game &
  \cellcolor[HTML]{FFF2CC}\tred{65.53} \\
  \hhline{~*{3}{-}}
 &
   &
  \gap &
  90.64 \\
 &
   &
  \cda &
  \cellcolor[HTML]{FFFFFF}{89.98} \\
 &
   &
  \tda &
  74.88 \\
  \hhline{~~*{2}{-}}
 &
  \multirow{-4}{*}{\dual{\dual{CUB-200, Stanford Cars,/ FGVC Aircraft}/(Fine-Grained tasks)}} &
  \cellcolor[HTML]{FFF2CC}\game &
  \cellcolor[HTML]{FFF2CC}\tred{63.39} \\
  \hhline{~*{3}{-}}
 &
   &
  \gap &
  73.25 \\
 &
   &
  \cda &
  71.94 \\
 &
   &
  \tda &
  42.37 \\
  \hhline{~~*{2}{-}}
\multirow{-12}{*}{\rot{\dual{\textit{Setting} 4/(difficult)}}} &
  \multirow{-4}{*}{ImageNet} &
  \cellcolor[HTML]{FFF2CC}\game &
  \cellcolor[HTML]{FFF2CC}\tred{27.51}\\
\hline

\end{tabular}%
}
\end{table}

%% file: tables/cd_method_comparison_voc.tex
\begin{table}[!t]
\caption{\textit{Setting 4 attack comparison when $\pertG$ is trained with \voc:} Perturbations created on test set of each task. $\bm{f}(\cdot)$: Res152.}
\vspace*{0.5\baselineskip}
\renewcommand{\aboverulesep}{0pt}
\renewcommand{\belowrulesep}{0pt}
\centering
\begin{subtable}[h]{\columnwidth}
\caption{Coarse-Grained task}
\resizebox{\columnwidth}{!}{
\begin{tabular}{ccccc}
\toprule
\rowcolor{red!10}
 & CIFAR10 & CIFAR100 & STL-10 & SVHN\\
 \cmidrule(l){2-5}
 \rowcolor{red!10}
 & \multicolumn{4}{c}{\textbf{All Victim Models from \cite{coarse2022task}}}\\
 \cmidrule(l){2-5} 
\rowcolor{red!10}
 \multirow{-3}{*}{\textbf{Method}} & \tgray{93.79\%} & \tgray{74.28\%} & \tgray{77.60\%}& \tgray{96.03\%} \\
\midrule
\gap & 92.94\% & 72.56\% & 74.33\% & 96.01\% \\
\cda & \tblue{91.97}\% & 72.18\% & \tblue{70.99}\% & \tblue{95.74}\% \\
\tda & 92.49\% & \tblue{70.80}\% & 73.31\% & 95.93\% \\
\midrule
\game & \tred{76.61}\% & \tred{47.51}\% & \tred{70.49}\% & \tred{88.27}\% \\
\bottomrule
\end{tabular}}
\label{tab:voc_coarse_attack}
\end{subtable}\\[0.9em]
\begin{subtable}[h]{\columnwidth}
\caption{Fine-Grained tasks}
\renewcommand{\aboverulesep}{0pt}
\renewcommand{\belowrulesep}{0pt}
\centering
\resizebox{\columnwidth}{!}{
\begin{tabular}{ccccccc}
\toprule
\rowcolor{red!10}
 & \multicolumn{2}{c}{\textbf{CUB-200-2011}} & \multicolumn{2}{c}{\textbf{Stanford Cars}} & \multicolumn{2}{c}{\textbf{FGVC Aircraft}}\\
 \cmidrule(l){2-7} 
 \rowcolor{red!10}
 & Res50 & SENet154 & Res50 & SENet154 & Res50 & SENet154\\
 \cmidrule(l){2-7} 
\rowcolor{red!10}
 \multirow{-3}{*}{\textbf{Method}} & \tgray{87.35\%} & \tgray{86.81\%} & \tgray{94.35\%}& \tgray{93.36\%} & \tgray{92.23\%} & \tgray{92.05\%}\\
\midrule
\gap & 86.24\% & 86.40\% & 93.79\% & 93.09\% & 
91.69\% & 91.78\% \\
\cda & 85.90\% & 86.11\% & 93.28\% & 92.69\% &
91.36\% & 91.90\% \\
\tda & \tblue{83.93}\% & \tblue{82.33}\% & \tblue{92.92}\% & \tblue{91.79}\% & \tblue{90.04}\% & \tblue{90.64}\%\\
\midrule
\game & \tred{59.34}\% & \tred{76.58}\% & \tred{77.35}\% & \tred{81.98}\% & \tred{73.78}\% & \tred{73.27}\% \\ 
\bottomrule
\end{tabular}}
\label{tab:voc_fine_attack}
\end{subtable}\\[0.9em]
\begin{subtable}[h]{\columnwidth}
\caption{ImageNet task (on ImageNet validation set (50k samples))}
\renewcommand{\aboverulesep}{0pt}
\renewcommand{\belowrulesep}{0pt}
\centering
\resizebox{\columnwidth}{!}{
\begin{tabular}{ccccccc}
\toprule
\rowcolor{red!10}
 & \multicolumn{6}{c}{\textbf{ImageNet Trained Victim Classifiers}} \\ \cmidrule(l){2-7} 
\rowcolor{red!10}
 & VGG16 & VGG19 & Res50 & Res152 & Dense121 & Dense169\\
 \cmidrule(l){2-7} 
\rowcolor{red!10}
 \multirow{-3}{*}{\textbf{Method}} & \tgray{70.15\%} & \tgray{70.95\%} & \tgray{74.60\%}& \tgray{77.34\%} & \tgray{74.21\%} & \tgray{75.74\%}\\
\midrule
\gap & 69.19\% & 70.23\% & 73.71\% & 76.62\% & 
73.36\% & 75.21\% \\
\cda & 68.20\% & 69.41\% & 72.67\% & 75.95\% &
72.93\% & 74.79\% \\
\tda & \tblue{65.60}\% & \tblue{66.28}\% & \tblue{70.47}\% & \tblue{74.35}\% & \tblue{70.11}\% & \tblue{72.62}\%\\
\midrule
\game & \tred{32.24}\% & \tred{35.05}\% & \tred{48.53}\% & \tred{50.54}\% & \tred{49.99}\% & \tred{54.37}\% \\ 
\bottomrule
\end{tabular}}
\label{tab:voc_imgnet_attack}
\end{subtable}
\label{tab:voc_sl}
\end{table}

%% file: tables/cd_method_comparison_coco.tex
\begin{table}[!t]
\caption{\textit{Setting 4 attack comparison when $\pertG$ is trained with \coco:} Perturbations created on test set of each task. $\bm{f}(\cdot)$: Dense169.}
\vspace*{0.5\baselineskip}
\renewcommand{\aboverulesep}{0pt}
\renewcommand{\belowrulesep}{0pt}
\centering
\begin{subtable}[h]{\columnwidth}
\caption{Coarse-Grained task}
\resizebox{\columnwidth}{!}{
\begin{tabular}{ccccc}
\toprule
\rowcolor{red!10}
 & CIFAR10 & CIFAR100 & STL-10 & SVHN\\
 \cmidrule(l){2-5} 
 \rowcolor{red!10}
 & \multicolumn{4}{c}{\textbf{All Victim Models from \cite{coarse2022task}}}\\
 \cmidrule(l){2-5} 
\rowcolor{red!10}
 \multirow{-3}{*}{\textbf{Method}} & \tgray{93.79\%} & \tgray{74.28\%} & \tgray{77.60\%}& \tgray{96.03\%} \\
\midrule
\gap & 93.12\% & 72.72\% & 74.78\% & 95.65\% \\
\cda & 90.77\% & 69.20\% & \tblue{70.31}\% & 95.79\% \\
\tda & \tblue{76.37}\% & \tblue{40.35}\% & 72.19\% & \tblue{92.67}\% \\
\midrule
\game & \tred{66.16}\% & \tred{35.12}\% & \tred{70.28}\% & \tred{90.56}\% \\ 
\bottomrule
\end{tabular}}
\label{tab:coco_coarse_attack}
\end{subtable}\\[0.9em]
\begin{subtable}[h]{\columnwidth}
\caption{Fine-Grained tasks}
\renewcommand{\aboverulesep}{0pt}
\renewcommand{\belowrulesep}{0pt}
\centering
\resizebox{\columnwidth}{!}{
\begin{tabular}{ccccccc}
\toprule
\rowcolor{red!10}
 & \multicolumn{2}{c}{\textbf{CUB-200-2011}} & \multicolumn{2}{c}{\textbf{Stanford Cars}} & \multicolumn{2}{c}{\textbf{FGVC Aircraft}}\\
 \cmidrule(l){2-7} 
 \rowcolor{red!10}
 & Res50 & SENet154 & Res50 & SENet154 & Res50 & SENet154\\
 \cmidrule(l){2-7} 
\rowcolor{red!10}
 \multirow{-3}{*}{\textbf{Method}} & \tgray{87.35\%} & \tgray{86.81\%} & \tgray{94.35\%}& \tgray{93.36\%} & \tgray{92.23\%} & \tgray{92.05\%}\\
\midrule
\gap & 86.69\% & 86.33\% & 94.12\% & 93.10\% & 
91.84\% & 91.78\% \\
\cda & 85.57\% & 86.04\% & 93.10\% & 92.71\% &
91.15\% & 91.30\% \\
\tda & \tblue{60.30}\% & \tred{70.04}\% & \tblue{76.21}\% & \tred{80.48}\% & \tblue{81.07}\% & \tblue{81.19}\%\\
\midrule
\game & \tred{22.25}\% & \tblue{74.77}\% & \tred{64.98}\% & \tblue{81.31}\% & \tred{60.37}\% & \tred{76.66}\% \\ 
\bottomrule
\end{tabular}}
\label{tab:coco_fine_attack}
\end{subtable}\\[0.9em]
\begin{subtable}[h]{\columnwidth}
\caption{ImageNet task (on ImageNet validation set (50k samples))}
\renewcommand{\aboverulesep}{0pt}
\renewcommand{\belowrulesep}{0pt}
\centering
\resizebox{\columnwidth}{!}{
\begin{tabular}{ccccccc}
\toprule
\rowcolor{red!10}
 & \multicolumn{6}{c}{\textbf{ImageNet Trained Victim Classifiers}} \\ \cmidrule(l){2-7} 
\rowcolor{red!10}
 & VGG16 & VGG19 & Res50 & Res152 & Dense121 & Dense169\\
 \cmidrule(l){2-7} 
\rowcolor{red!10}
 \multirow{-3}{*}{\textbf{Method}} & \tgray{70.15\%} & \tgray{70.95\%} & \tgray{74.60\%}& \tgray{77.34\%} & \tgray{74.21\%} & \tgray{75.74\%}\\
\midrule
\gap & 69.32\% & 70.39\% & 73.89\% & 76.75\% & 
73.75\% & 75.38\% \\
\cda & 67.24\% & 68.45\% & 72.17\% & 75.69\% &
73.12\% & 74.96\% \\
\tda & \tblue{31.59}\% & \tblue{33.11}\% & \tblue{45.74}\% & \tblue{58.15}\% & \tblue{46.11}\% & \tblue{39.49}\% \\
\midrule
\game & \tred{20.60}\% & \tred{23.60}\% & \tred{30.42}\% & \tred{37.07}\% & \tred{29.50}\% & \tred{23.88}\% \\ 
\bottomrule
\end{tabular}}
\label{tab:coco_imgnet_attack}
\end{subtable}
\label{tab:coco_sl}
\end{table}

%% file: tables/wb_bb_method_comparison_voc.tex
\begin{table*}[!t]
    \renewcommand{\aboverulesep}{0pt}
    \renewcommand{\belowrulesep}{0pt}
    \setlength{\tabcolsep}{10pt}
    \caption{\textit{Generative Attack Comparison when $\pertG$ is trained with \voc:} \colorbox{gray!15}{Gray} colored cells represent the \textit{Setting} 1 attacks. $\bm{f}(\cdot)$ in both \cref{tab:wb_voc} and \cref{tab:cd_voc} are pre-trained on \voc.}
    \centering
    \vspace{0.75\baselineskip}
\begin{subtable}[h]{\columnwidth}
    \caption{\textit{Setting} 1 and \textit{Setting} 2 attacks}
    \centering
    \resizebox{\columnwidth}{!}{
    \begin{tabular}{ccccc}
    \toprule
    \rowcolor{red!10}
     & & \multicolumn{3}{c}{\textbf{\voc Trained Victim Models}} \\
     \cmidrule(l){3-5}
    \rowcolor{red!10}
     & & ~\textbf{Res152}~ & ~\textbf{VGG19}~ &  \textbf{Dense169}\\
    \cmidrule(l){3-5}
    \rowcolor{red!10}
    \multirow{-3}{*}{$\bm{f}(\cdot)$} & \multirow{-3}{*}{\textbf{Method}}  & \tgray{83.12\%} & \tgray{83.18\%} & \tgray{83.73\%} \\
    \midrule 
     & \gap & \cellcolor[HTML]{EFEFEF} 58.78\% & 48.52\% & 61.31\% \\
     & \cda & \cellcolor[HTML]{EFEFEF} 58.62\% & 48.69\% & \tblue{60.93}\% \\
     & \tda & \cellcolor[HTML]{EFEFEF} \tblue{58.45}\% & \tblue{48.19}\% & 61.16\% \\
    \cmidrule(l){2-5}
    \multirow{-4}{*}{\rot{{\textbf{Res152}}}} & \game & \cellcolor[HTML]{EFEFEF} \tred{57.22}\% & \tred{46.07}\% & \tred{59.63}\% \\
    \midrule
     & \gap & 58.88\% & \cellcolor[HTML]{EFEFEF} 45.60\% & 61.32\% \\
     & \cda & 58.18\% & \cellcolor[HTML]{EFEFEF} 45.26\% & 60.73\% \\
     & \tda & \tred{57.47}\% & \cellcolor[HTML]{EFEFEF} \tred{42.61}\% & \tred{59.39}\% \\
    \cmidrule(l){2-5}
    \multirow{-4}{*}{\rot{{\textbf{VGG19}}}} & \game & \tblue{57.84}\% & \cellcolor[HTML]{EFEFEF} \tblue{42.62}\% & \tblue{59.66}\% \\
    \midrule
     & \gap & 58.83\% & 48.58\% & \cellcolor[HTML]{EFEFEF} 61.29\% \\
     & \cda & 58.39\% & 48.25\% & \cellcolor[HTML]{EFEFEF} 60.48\% \\
     & \tda & \tblue{58.04}\% & \tblue{47.66}\% & \cellcolor[HTML]{EFEFEF} \tblue{60.12}\% \\
    \cmidrule(l){2-5}
    \multirow{-4}{*}{\rot{\small{\textbf{Dense169}}}} & \game & \tred{57.21}\% & \tred{45.82}\% & \cellcolor[HTML]{EFEFEF} \tred{58.23}\% \\
    \bottomrule
    \end{tabular}}
    \label{tab:wb_voc}
\end{subtable}%
\hfill
\begin{subtable}[h]{\columnwidth}
    \caption{\textit{Setting} 3 attacks}
    \centering
    \resizebox{0.995\columnwidth}{!}{
    \begin{tabular}{ccccc}
    \toprule
    \rowcolor{red!10}
     & & \multicolumn{3}{c}{\textbf{\coco Trained Victim Models}} \\
     \cmidrule(l){3-5}
    \rowcolor{red!10}
     & & ~\textbf{Res152}~ & ~\textbf{VGG19}~ &  \textbf{Dense169}\\
    \cmidrule(l){3-5}
    \rowcolor{red!10}
    \multirow{-3}{*}{$\bm{f}(\cdot)$} & \multirow{-3}{*}{\textbf{Method}}  & \tgray{67.95\%} & \tgray{66.49\%} & \tgray{67.60\%} \\ 
    \midrule 
     & \gap &  {44.91}\% & 34.70\% & 44.15\% \\
     & \cda &  44.99\% & 34.89\% & 44.35\% \\
     & \tda &  \tblue{44.45}\% & \tblue{34.46}\% & \tblue{43.88}\% \\
    \cmidrule(l){2-5}
    \multirow{-4}{*}{\rot{\textbf{Res152}}} & \game & \tred{42.36}\% & \tred{32.16\%} & \tred{42.37}\% \\
    \midrule
     & \gap & 45.02\% &  31.10\% & 44.14\% \\
     & \cda & 43.41\% &  30.94\% & 43.31\% \\
     & \tda & \tblue{43.22}\% &  \tred{27.74}\% & \tred{42.23}\% \\
    \cmidrule(l){2-5}
    \multirow{-4}{*}{\rot{\textbf{VGG19}}} & \game & \tred{43.12}\% &  \tblue{28.31}\% & \tblue{42.54}\% \\
    \midrule
     & \gap & 44.88\% & 34.72\% &  44.12\% \\
     & \cda & 44.50\% & 34.42\% &  43.82\% \\
     & \tda & \tblue{44.21}\% & \tblue{34.30}\% & \tblue{43.58}\% \\
    \cmidrule(l){2-5}
    \multirow{-4}{*}{\rot{\small{\textbf{Dense169}}}} & \game & \tred{43.09\%} & \tred{32.40}\% &  \tred{41.86}\% \\
    \bottomrule
    \end{tabular}}
    \label{tab:cd_voc}
\end{subtable}
\label{tab:main_voc}
\vspace{-0.15\baselineskip}
\end{table*}

%% file: figures/ablation/ablation_analysis.tex
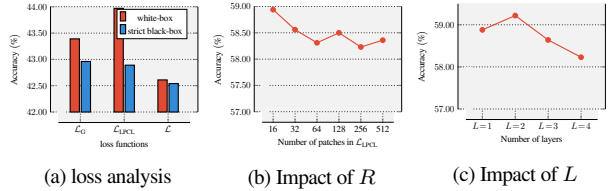
\begin{figure}[!ht]
    \centering
    \vspace{-0.75\baselineskip}
    \begin{subfigure}{.325\columnwidth}
        \input{figures/ablation/loss_ablation}
        \caption{loss analysis}
        \label{fig:loss_abl}
    \end{subfigure}%
    \begin{subfigure}{.325\columnwidth}
        \input{figures/ablation/num_patches}
        \caption{Impact of $R$}
        \label{fig:patches}
    \end{subfigure}%
    \begin{subfigure}{.325\columnwidth}
        \input{figures/ablation/num_layers}
        \caption{Impact of $L$}
        \label{fig:layers}
    \end{subfigure}
    \vspace{0.5\baselineskip}
    \caption{\textit{Ablation analysis of \game:} \cref{fig:loss_abl}: $\pertG$ trained on \voc against Res152, strict black-box attacks on \coco; \cref{fig:patches}, \cref{fig:layers}: $\pertG$ trained on \voc against Dense169 for all cases; perturbation bound was set $\ell_\infty\leq10$.}
    \label{fig:abl_analysis}
\end{figure}

%% file: figures/ablation/loss_ablation.tex
\pgfplotsset{
   every axis/.append style = {
      line width = 1pt,
      tick style = {line width=1pt}
   }
}
\begin{tikzpicture}[scale = 0.35, transform shape]
\pgfplotsset{
    height=4cm, width=5cm,
    grid = major,
    grid = both,
    grid style = {
    dash pattern = on 0.05mm off 1mm,
    line cap = round,
    black,
    line width = 0.75pt},
    scale only axis,
    y tick label style={
        /pgf/number format/.cd,
        fixed,
        fixed zerofill,
        precision=2,
        /tikz/.cd
    }
}
\begin{axis}[
    xmin=0, xmax=6,
    xshift=-0.3cm,
    hide x axis,
    axis y line*=left,
    ymin=42, ymax=44,
    ylabel={Accuracy \big{(}\%\big{)}},
    axis background/.style={fill=gray!10}
]
\end{axis}
\begin{axis}[
    height=2cm, yshift=-0.4cm,
    xmin=0, xmax=6,
    ymin=42, ymax=44,
    axis x line*=bottom,
    hide y axis,
    xtick = {1, 3, 5},
    xticklabels = {$\mse$, $\ccl$, $\mathcal{L}$},
    xlabel={loss functions},
    axis background/.style={fill=gray!10}
]
\end{axis}
\begin{axis}[
    xmin=0, xmax=6,
    ymin=42, ymax=44,
    yticklabels = {},
    xshift=0.3cm,
    hide x axis,
    axis y line*=right,
    axis background/.style={fill=gray!10}
]
\end{axis}
\begin{axis}[
    ybar,
    xmin=0, xmax=6,
    ymin=42, ymax=44,
    hide x axis,
    hide y axis,
    legend image code/.code={
        \draw [#1] (0cm,-0.1cm) rectangle (0.2cm,0.25cm); },
]
]
    \addplot[very thick, draw = black, fill = r1-color]
    coordinates {(1,43.39)(3,43.97)(5,42.61)};
    \addplot[very thick, draw = black, fill = comment-color] 
    coordinates {(1,42.96)(3,42.89)(5,42.54)};
    \legend{white-box, strict black-box}
\end{axis}
\end{tikzpicture}

%% file: figures/ablation/num_patches.tex
\pgfplotsset{
   every axis/.append style = {
      line width = 1pt,
      tick style = {line width=1pt}
   }
}
\begin{tikzpicture}[scale = 0.35, transform shape]
    \pgfplotsset{
        height=4cm, width=5cm,
        grid = major,
        grid = both,
        grid style = {
        dash pattern = on 0.05mm off 1mm,
        line cap = round,
        black,
        line width = 0.75pt},
        scale only axis,
        y tick label style={
        /pgf/number format/.cd,
        fixed,
        fixed zerofill,
        precision=2,
        /tikz/.cd
    }
    }
    \begin{axis}[
        xmin=0, xmax=12,
        ymin=57, ymax=59,
        xshift=-0.3cm,
        hide x axis,
        axis y line*=left,
        ylabel={Accuracy \big{(}\%\big{)}},
        axis background/.style={fill=gray!10}
    ]
    \end{axis}
    \begin{axis}[
        height=2cm, yshift=-0.4cm,
        xmin=0, xmax=12,
        ymin=57, ymax=59,
        axis x line*=bottom,
        hide y axis,
        xtick = {1, 3, 5, 7, 9, 11},
        xticklabels = {16, 32, 64, 128, 256, 512},
        xlabel={Number of patches in $\ccl$},
        axis background/.style={fill=gray!10}
    ]
    \end{axis}
    \begin{axis}[
        xmin=0, xmax=12,
        ymin=57, ymax=59,
        yticklabels = {},
        xshift=0.3cm,
        hide x axis,
        axis y line*=right,
        axis background/.style={fill=gray!10}
    ]
    \end{axis}
    \begin{axis}[
        xmin=0, xmax=12,
        ymin=57, ymax=59,
        hide x axis,
        hide y axis,
    ]
        \addplot[very thick, r1-color, mark=*, mark options={solid, scale=1.25, fill=r1-color}] 
        coordinates {(1,58.94)(3,58.56)(5,58.31)(7,58.50)(9,58.23)(11,58.36)};
    \end{axis}
    
\end{tikzpicture}

%% file: figures/ablation/num_layers.tex
\pgfplotsset{
   every axis/.append style = {
      line width = 1pt,
      tick style = {line width=1pt}
   }
}
\begin{tikzpicture}[scale = 0.35, transform shape]
    \pgfplotsset{
        height=4cm, width=5cm,
        grid = major,
        grid = both,
        grid style = {
        dash pattern = on 0.05mm off 1mm,
        line cap = round,
        black,
        line width = 0.75pt},
        scale only axis,
        y tick label style={
        /pgf/number format/.cd,
        fixed,
        fixed zerofill,
        precision=2,
        /tikz/.cd
    }
    }
    \begin{axis}[
        xmin=0, xmax=8,
        xshift=-0.3cm,
        hide x axis,
        axis y line*=left,
        ymin=57, ymax=59.5,
        ylabel={Accuracy \big{(}\%\big{)}},
        axis background/.style={fill=gray!10}
    ]
    \end{axis}
    \begin{axis}[
        height=2cm, yshift=-0.4cm,
        xmin=0, xmax=8,
        ymin=57, ymax=59.5,
        axis x line*=bottom,
        hide y axis,
        xtick = {1, 3, 5, 7},
        xticklabels = {$L=1$, $L=2$, $L=3$, $L=4$},
        xlabel={Number of layers},
        axis background/.style={fill=gray!10}
    ]
    \end{axis}
    \begin{axis}[
        xmin=0, xmax=8,
        ymin=57, ymax=59.5,
        yticklabels = {},
        xshift=0.3cm,
        hide x axis,
        axis y line*=right,
        axis background/.style={fill=gray!10}
    ]
    \end{axis}
    \begin{axis}[
        xmin=0, xmax=8,
        ymin=57, ymax=59.5,
        hide x axis,
        hide y axis,
    ]
        \addplot[very thick, r1-color, mark=*, mark options={solid, scale=1.25, fill=r1-color}] 
        coordinates {(1,58.88)(3,59.22)(5,58.64)(7,58.23)};
    \end{axis}
\end{tikzpicture}

%% file: figures/qual_result.tex
\begin{figure*}[!ht]
  \centering
  \includegraphics[width=\textwidth]{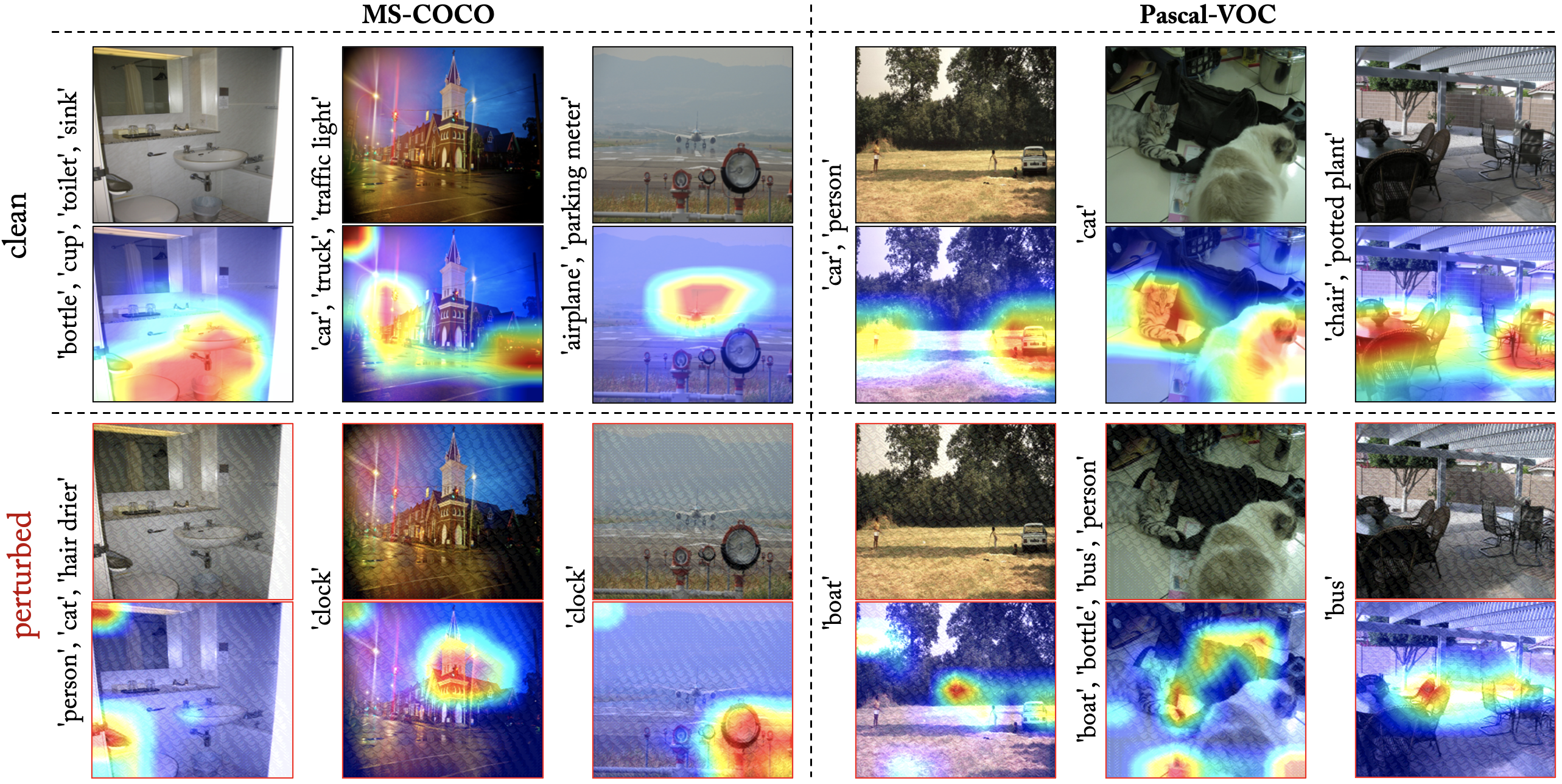}
  \caption{\textit{Illustration of perturbed images and attention shift:} Row 1: clean images, Row 2: CAM \cite{zhou2016learning} attention map on clean images, Row 3: perturbed images ($\ell_\infty\leq10$), Row 4: CAM \cite{zhou2016learning} attention map on perturbed images. $\pertG$ was trained against Res152 for both datasets, examples are visualized on test sets with attention maps extracted from Res152.} 
  \label{fig:qual_result}
\end{figure*}

%% file: sections/5_conclusion.tex
\section{Conclusion}
In this paper, we tackle a novel problem of altering the decisions of victim classifiers by learning to create perturbations on multi-object images. To this end, we proposed a novel generative adversarial attack (\game) framework that trains the perturbation generators by exploiting the local differences in multi-object image features. \game achieves high attack rates both in white-box and different practical black-box settings. For example, when we learn to craft perturbations on \voc and create black-box attack on ImageNet, \game outperforms existing attacks by $\sim$25\% points. In our future work, we will explore the case of black-box multi-object targeted attacks for multi-object images, as well as video generative models \cite{Aich_2020_CVPR, gupta2020alanet} for adversarial attacks on video classifiers.

\textbf{Acknowledgement.} This material is based upon work supported by the Defense Advanced Research Projects Agency (DARPA) under Agreement No. HR00112090096. 

%% file: supp_material/1_add_exps.tex
\paragraph{Baselines.} We use three state-of-the-art generative attack methods (\gap, \cda, and \tda) as our baselines. We change the cross-entropy loss in \gap and \cda with binary cross-entropy loss to adapt it for multi-object surrogate classifier.
\newline
\paragraph{Implementation Details.} Following \cite{naseer2019cross}, we use the ResNet architecture introduced in \cite{johnson2016perceptual} as the generator network for $\pertG$, except we replace the ReLU \cite{nair2010rectified} activation with Fused Leaky ReLU \cite{karras2019style} activation to stabilize training (negative slope = 0.2, scale set = $\sqrt{2}$). We use Adam optimizer \cite{kingma2014adam} with a learning rate $0.0001$, batch size 32, and exponential decay rates between 0.5 and 0.999. All images are resized to $224\times 224$ and normalized with mean and standard deviation before feeding to generator. Further, similar to gaussian smoothing in \cda, in order to make the perturbations more transferable, we clamp the perturbed image between 0 to 1. This clamping trick helps in increasing the transferability of the perturbations. For fair comparison, we apply this strategy to all attacks. Perturbation generators are trained for 20 epochs. We use PyTorch \cite{paszke2019pytorch} in all experiments. Training time is 1 hr for Pascal-VOC dataset and 10 hrs for MS-COCO dataset on two NVIDIA GeForce RTX 3090 GPUs. For all experiments, number of patches for $\ccl$ is set to 128. 
\newline
\paragraph{White-box, black-box, and strict black-box attacks.} We sanalyze white-box and black-box attack (attack in same distribution as adversary) performance of \game in \cref{tab:wb_coco}. This attack tests the strength of perturbations on the same type of task (\ie multi-object classification) as training. Our proposed method shows a stronger attack than \gap and \cda. In comparison to \tda, our attack shows comparable performances (for cases where \tda does better, the difference is very small) in most cases even though we do not need to manually choose a specific layer for each classifier to train the perturbation generator. Choosing a particular mid-layer for every classifier does not always guarantee better transferability of perturbations. A similar observation can be made in \cref{tab:cd_coco}.

\input{tables/wb_bb_method_comparison_coco}

%% file: tables/wb_bb_method_comparison_coco.tex
\begin{table*}[!ht]
    \renewcommand{\aboverulesep}{0pt}
    \renewcommand{\belowrulesep}{0pt}
    \setlength{\tabcolsep}{10pt}
    \caption{\textit{Generative Attack Comparison when $\pertG$ is trained with \coco:} \colorbox{gray!15}{Gray} colored cells represent the white-box attacks. $\bm{f}(\cdot)$ in both \cref{tab:wb_coco} and \cref{tab:cd_coco} are pre-trained on \coco.}
    \vspace{\baselineskip}
    \centering
\begin{subtable}[h]{0.495\columnwidth}
    \caption{\textit{Setting} 1 and \textit{Setting} 2}
    \centering
    \resizebox{\columnwidth}{!}{
    \begin{tabular}{ccccc}
    \toprule
    \rowcolor{red!10}
     & & \multicolumn{3}{c}{\textbf{\coco Trained Victim Models}} \\
     \cmidrule(l){3-5}
    \rowcolor{red!10}
     & & ~\textbf{Res152}~ & ~\textbf{VGG19}~ &  \textbf{Dense169}\\
    \cmidrule(l){3-5}
    \rowcolor{red!10}
    \multirow{-3}{*}{$\bm{f}(\cdot)$} & \multirow{-3}{*}{\textbf{Method}}  & \tgray{67.95\%} & \tgray{66.49\%} & \tgray{67.60\%} \\
    \midrule 
     & \gap & \cellcolor[HTML]{EFEFEF} 44.98\% & 34.61\% & 43.91\% \\
     & \cda & \cellcolor[HTML]{EFEFEF} 45.00\% & 34.91\% & {44.36}\% \\
     & \tda & \cellcolor[HTML]{EFEFEF} \tred{39.60}\% & \tred{29.41}\% & \tred{39.66}\% \\
    \cmidrule(l){2-5}
    \multirow{-4}{*}{\rot{\textbf{Res152}}} & \game & \cellcolor[HTML]{EFEFEF} \tblue{41.02}\% & \tblue{30.18}\% & \tblue{40.33}\% \\
    \midrule
     & \gap & 44.83\% & \cellcolor[HTML]{EFEFEF} 34.67\% & 44.10\% \\
     & \cda & 44.41\% & \cellcolor[HTML]{EFEFEF} 30.63\% & 43.53\% \\
     & \tda & \tred{39.81}\% & \cellcolor[HTML]{EFEFEF} \tred{23.04}\% & \tred{38.96}\% \\
    \cmidrule(l){2-5}
    \multirow{-4}{*}{\rot{\textbf{VGG19}}} & \game & \tblue{40.09}\% & \cellcolor[HTML]{EFEFEF} \tblue{24.23}\% & \tblue{39.11}\% \\
    \midrule
     & \gap & 44.55\% & 34.28\% & \cellcolor[HTML]{EFEFEF} 43.61\% \\
     & \cda & 44.92\% & 34.86\% & \cellcolor[HTML]{EFEFEF} 44.24\% \\
     & \tda & \tblue{42.69}\% & \tblue{31.96}\% & \cellcolor[HTML]{EFEFEF} \tblue{40.30}\% \\
    \cmidrule(l){2-5}
    \multirow{-4}{*}{\rot{\small{\textbf{Dense169}}}} & \game & \tred{41.96}\% & \tred{30.19}\% & \cellcolor[HTML]{EFEFEF} \tred{39.47}\% \\
    \bottomrule
    \end{tabular}}
    \label{tab:wb_coco}
\end{subtable}%
\hfill
\begin{subtable}[!t]{0.495\columnwidth}
    \caption{\textit{Setting} 3}
    \centering
    \resizebox{\columnwidth}{!}{
    \begin{tabular}{ccccc}
    \toprule
    \rowcolor{red!10}
     & & \multicolumn{3}{c}{\textbf{\voc Trained Victim Models}} \\
     \cmidrule(l){3-5}
    \rowcolor{red!10}
     & & ~\textbf{Res152}~ & ~\textbf{VGG19}~ &  \textbf{Dense169}\\
    \cmidrule(l){3-5}
    \rowcolor{red!10}
    \multirow{-3}{*}{$\bm{f}(\cdot)$} & \multirow{-3}{*}{\textbf{Method}}  & \tgray{83.12\%} & \tgray{83.18\%} & \tgray{83.73\%} \\ 
    \midrule 
     & \gap &  {58.80}\% & 48.67\% & 60.80\% \\
     & \cda &  58.67\% & 48.66\% & 60.92\% \\
     & \tda &  \tred{54.05}\% & \tred{43.29}\% & \tred{57.55}\% \\
    \cmidrule(l){2-5}
    \multirow{-4}{*}{\rot{\textbf{Res152}}} & \game & \tblue{55.76}\% & \tblue{43.88\%} & \tblue{58.15}\% \\
    \midrule
     & \gap & 59.09\% &  48.61\% & 61.17\% \\
     & \cda & 58.44\% &  45.20\% & 60.35\% \\
     & \tda & \tred{55.06}\% &  \tred{38.41}\% & \tred{57.49}\% \\
    \cmidrule(l){2-5}
    \multirow{-4}{*}{\rot{\textbf{VGG19}}} & \game & \tblue{55.24}\% &  \tblue{40.71}\% & \tblue{58.10}\% \\
    \midrule
     & \gap & 58.45\% & 48.18\% &  60.47\% \\
     & \cda & 58.69\% & 48.68\% &  61.04\% \\
     & \tda & \tblue{56.65}\% & \tblue{45.52}\% & \tblue{58.54}\% \\
    \cmidrule(l){2-5}
    \multirow{-4}{*}{\rot{\small{\textbf{Dense169}}}} & \game & \tred{56.08\%} & \tred{43.46}\% &  \tred{57.23}\%\\
    \bottomrule
    \end{tabular}}
    \label{tab:cd_coco}
\end{subtable}
\label{tab:main_coco}
\end{table*}